\RequirePackage{fix-cm}
\documentclass[smallextended]{svjour3}       
\smartqed  
\usepackage{graphicx}
\graphicspath{{figs/}}
\usepackage[numbers]{natbib} 
\usepackage{amssymb,amsmath}
\usepackage[boxed]{algorithm2e}
\usepackage{subfigure}
\usepackage{url}
\newcommand{\Prob} {\ensuremath \mathbf{P}  }

\begin{document}

\title{Modeling Curiosity in a Mobile Robot for Long-Term Autonomous Exploration and Monitoring 
}


\author{Yogesh Girdhar         \and
        Gregory Dudek 
}


\institute{Yogesh Girdhar \at
              Applied Ocean Physics \& Engineering, Woods Hole Oceanographic Institution, Woods Hole, MA 02543, USA.\\
              \email{yogi@whoi.edu}           
           \and
           Gregory Dudek \at
              School of Computer Science\\ McGill University, Montreal, QC H3A0E9, Canada.\\
              \email{dudek@cim.mcgill.ca}
}

\date{Received: date / Accepted: date}

\maketitle
\begin{abstract}
This paper presents a novel approach to modeling curiosity in a mobile robot, which is useful for monitoring and adaptive data collection tasks, especially in the context of long term autonomous missions where pre-programmed missions are likely to have limited utility. We use a realtime topic modeling technique to build a semantic perception model of the environment, using which, we plan a path through the locations in the world with high semantic information content. The life-long learning behavior of the proposed perception model makes it suitable for long-term exploration missions. We validate the approach using simulated exploration experiments using aerial and underwater data, and demonstrate an implementation on the Aqua underwater robot in a variety of scenarios. We find that the proposed exploration paths that are biased towards locations with high topic perplexity, produce better terrain models with high discriminative power. Moreover, we show that the proposed algorithm implemented on Aqua robot is able to do tasks such as coral reef inspection, diver following, and sea floor exploration, without any prior training or preparation.

\end{abstract}

\section{Introduction}
Gaining knowledge about our environment is a never-ending quest for humanity. Direct exploration by humans although tempting, puts strong limitations on what can be explored due to the physical limitations of the human body. Fortunately, through the use of robotics, we can continue this tradition of exploration without putting human lives at risk. 

Use of autonomous robots is essential for space and ocean exploration, where there are strong communication bottlenecks that do not allow direct remote control of the vehicles \cite{Bellingham2007}. However, such exploration missions, which are inherently long-term, necessitate autonomy beyond low level navigational control. To maximize the utility of a mission in terms of information content of the collected data, there is a need for high level understanding of the environment in real time, which can then be used to adaptively plan the robot path.

A common approach for autonomous collection of environment data is to use space filling paths through the environment. This approach, although simple, is however not ideal. The amount of information collected that is associated with the different spatial phenomena, is proportional to the spatial area covered by them. Underwater, this might mean that most of the collected data only contains uninteresting observations of sand or rocks, and very occasionally we might have a few samples with something interesting such as thermal vents, marine life, or archeological sites. A better strategy for collecting data is to have the robot behave like an explorer, or a vacationing tourist, moving swiftly over regions with familiar sights while paying much more attention, i.e., collecting more data when something novel or interesting is in view. In this paper we describe such a techniques, and demonstrate its functioning on an underwater robot.

\begin{figure}[t]
\begin{center}
\includegraphics[width=0.4\columnwidth]{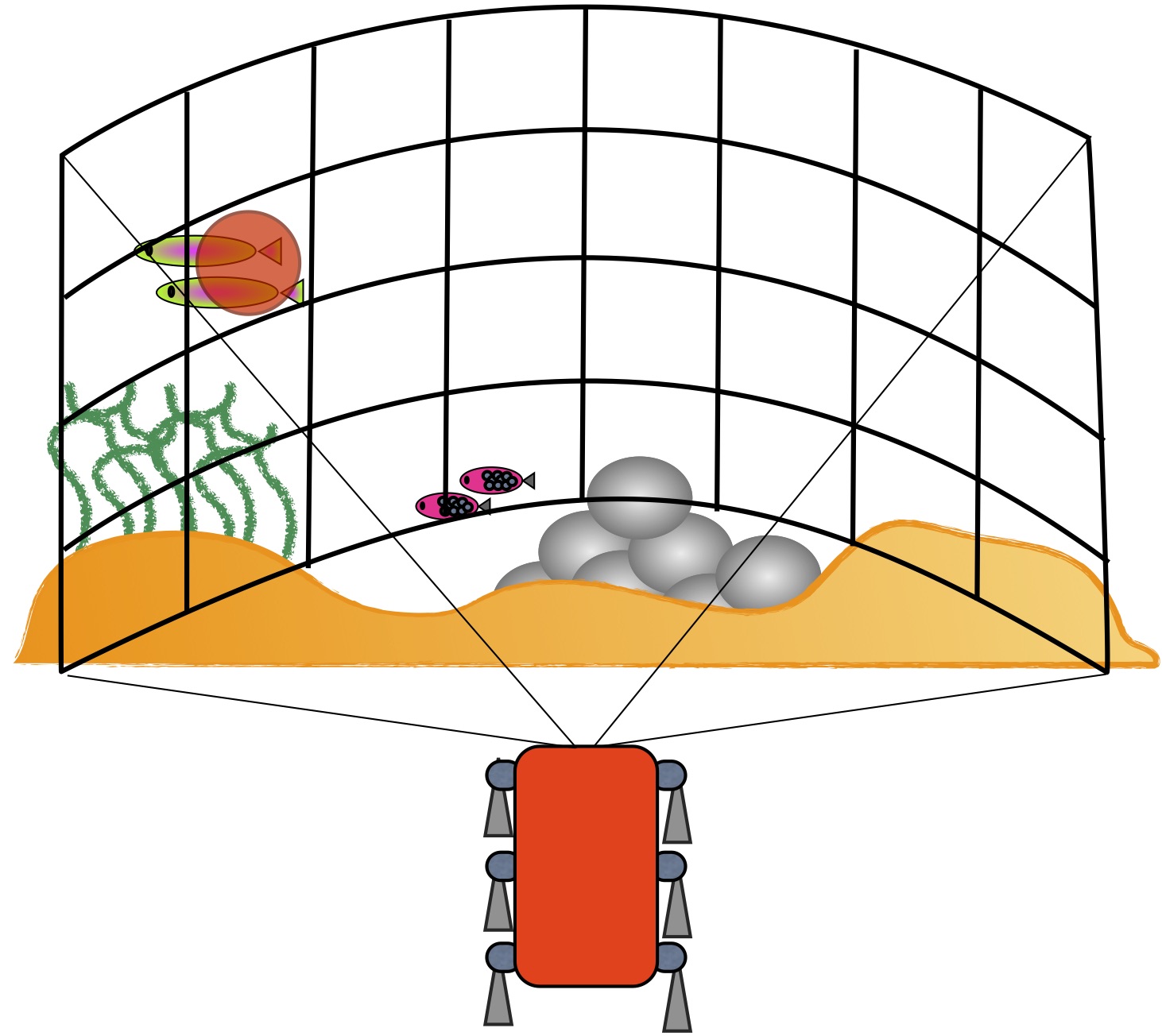}
\includegraphics[width=0.4\columnwidth]{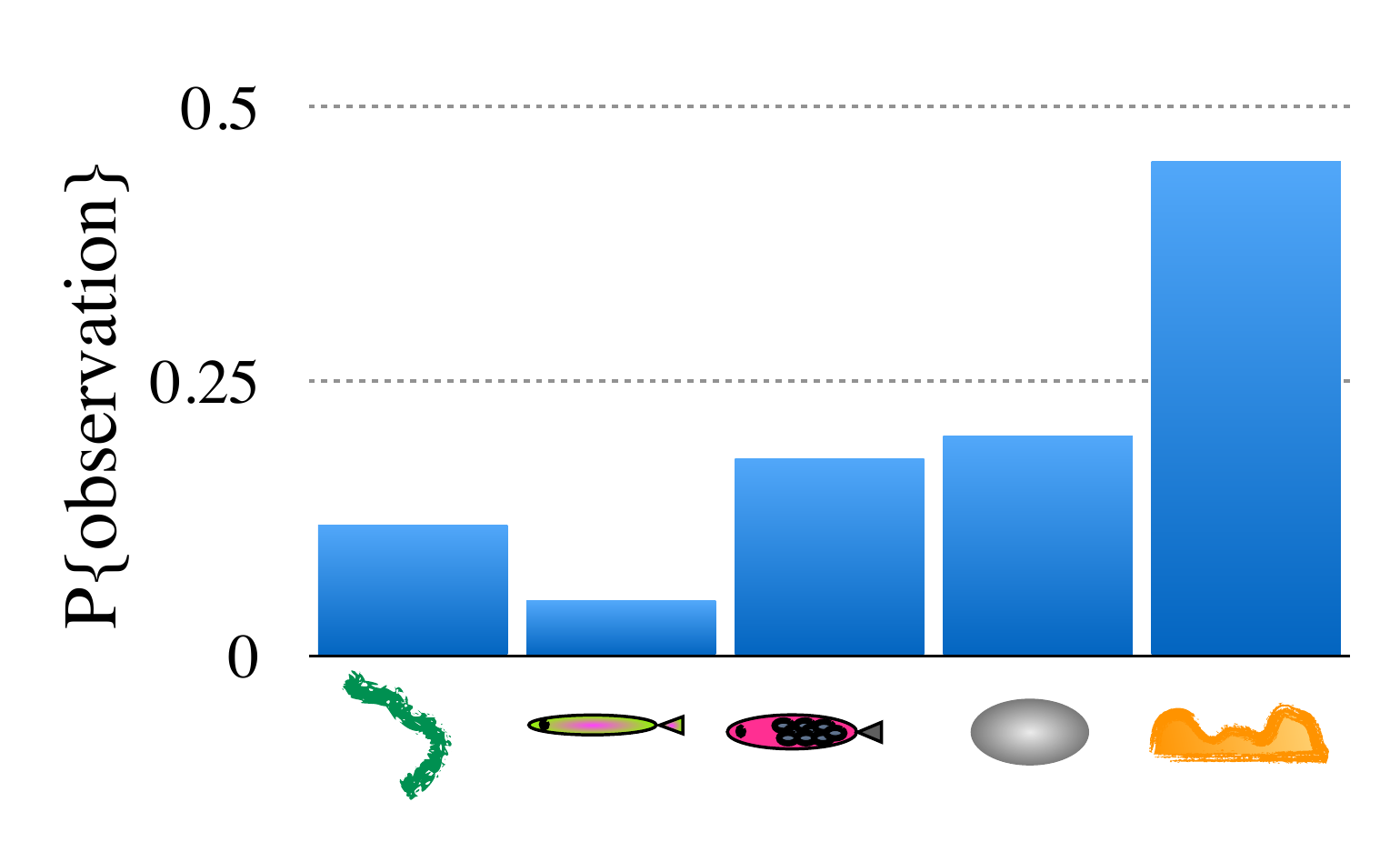}

\caption{An illustrated example of a scenario that demonstrates the proposed exploration strategy. The robot explores the environment while building an unsupervised topic model of its experience, which it then uses to find surprising and novel observations (highlighted by red circle). Topic correspond to various semantic scene constructs such as sand, rocks, coral and fish. Robot biases its exploration path in the direction of these surprising or novel observations to collect more data about them. Given the distribution of topics observed thus far (shown in the plot on the right), the most interesting observation in robot's view is the long fish.}\label{fig:overview}
\end{center}
\end{figure}

Our approach to identifying what is interesting is to first learn a generative visual model of the environment. Then, given this visual model we quantify the interestingness of an observed image sample by computing its perplexity score, i.e., how much uncertainty does the model have in describing what it has observed.  We use realtime online topic modeling (ROST) \cite{Girdhar2013IJRR}, to learn a constantly evolving visual model of the environment. ROST models the underlying cause of the observations made by the robot with a latent variable (called topic), which is representative of different kinds of terrains or other visual constructs in the scene. Topic modeling techniques have been shown to produce semantic labeling of text \cite{Blei:2003} and images \cite{Bourgault2002}, including satellite maps~\cite{Lienou2010}.




At each time step, we add the observations from the  current location to the topic model, and compute the perplexity of the observations from the neighboring observable locations. This perplexity score, along with a repulsive potential from previously visited locations, is then used to bias the probability of next step in the path. Since observations with high perplexity have high information gain, we claim that this approach would results in faster learning of the terrain topic model, which would imply shorter exploration paths for the same accuracy in predicting terrain labels for unseen regions. 

\begin{figure}[]
\begin{center}
\includegraphics[width=0.8\columnwidth]{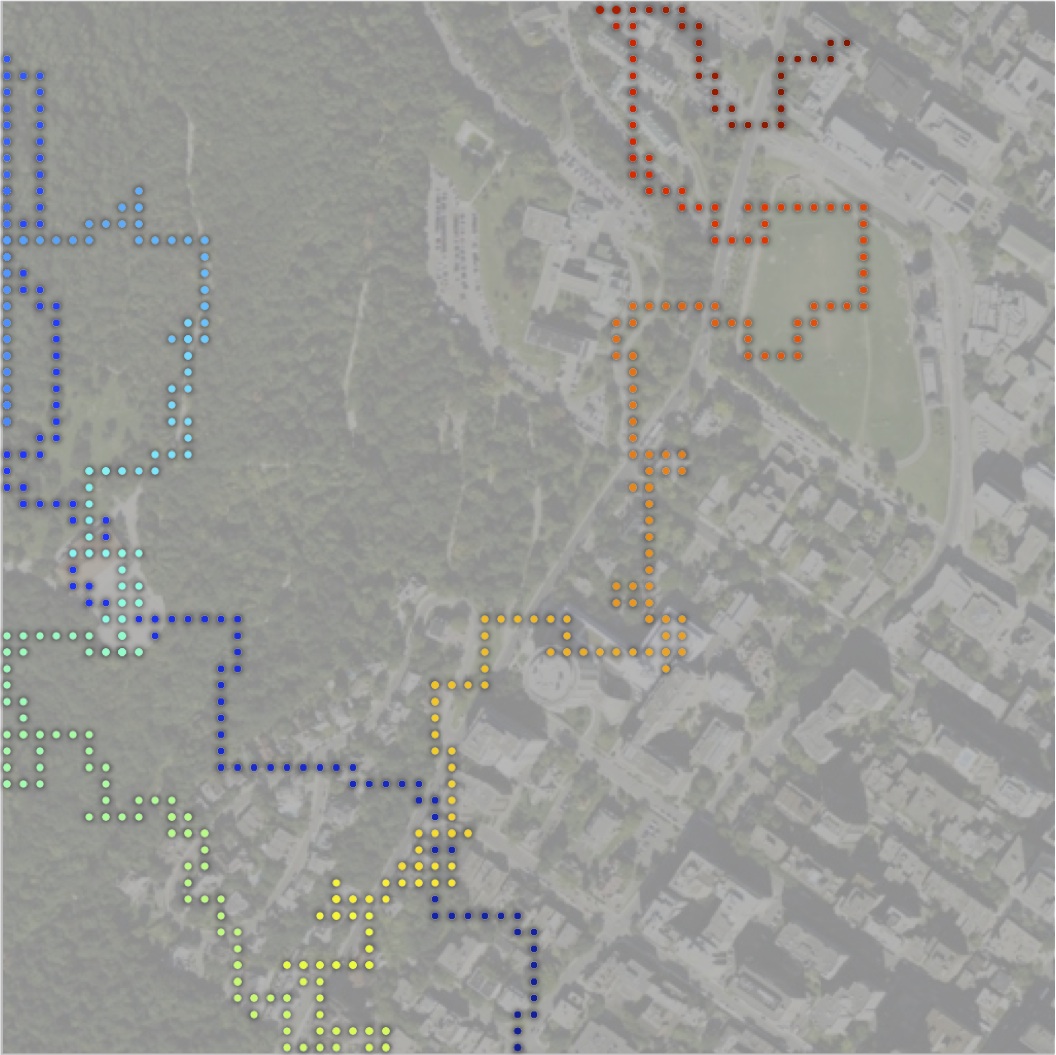}\\
\includegraphics[width=0.8\columnwidth]{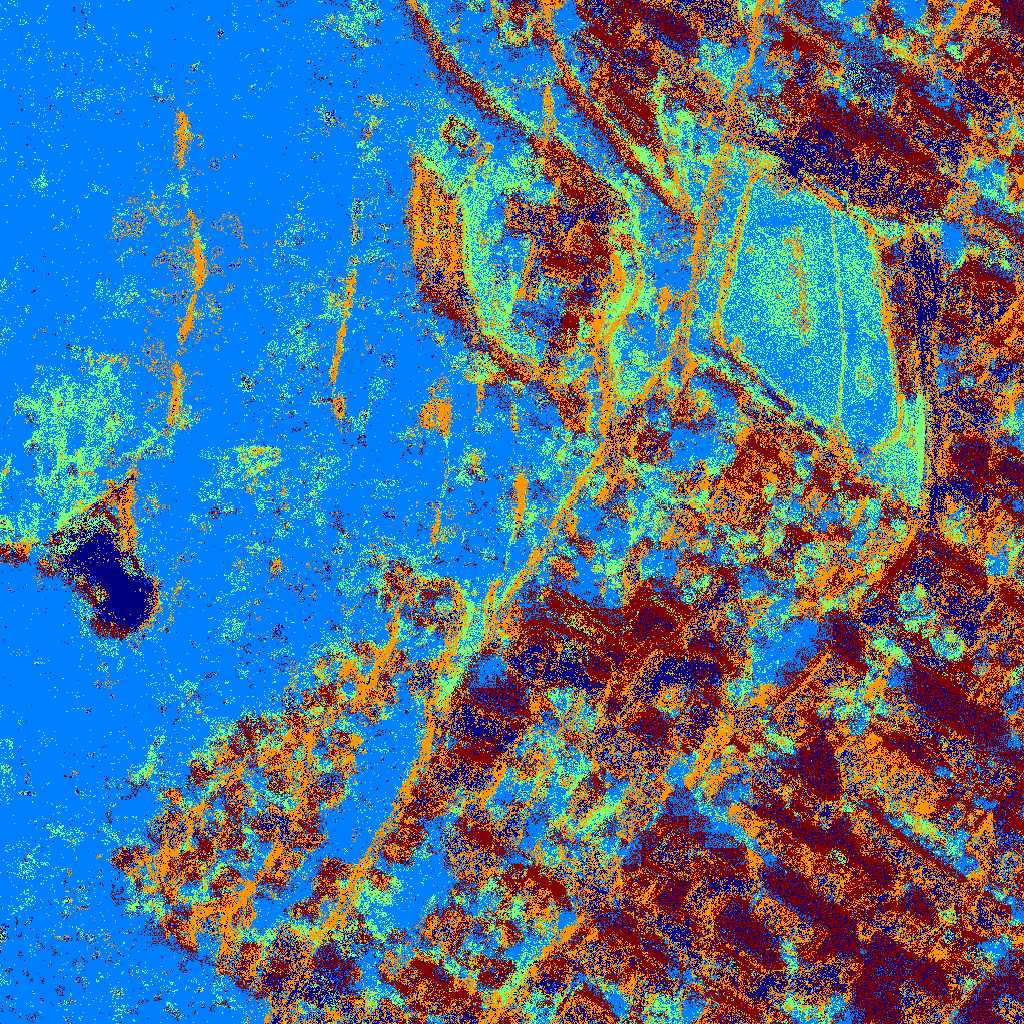}
\caption{Example of an exploratory path (top) produced by the proposed technique on a satellite map. The path begins in Blue, and ends in Red. Output of this exploration is a terrain model, which when applied to the observation from entire map produces terrain (topic) label for every location(bottom). Different colors represent different terrains (topic labels). }
\label{fig:eg_exploration}
\end{center}
\end{figure}

Figure \ref{fig:eg_exploration} (top) shows example of such an exploration path overlaid on top of an aerial view map. We see that the exploration path, which starts with blue, and ends in red, has in the beginning no preference over what is interesting, and hence is somewhat straight due to the repulsive potential from previously visited location. However after some time, in the cyan region of the path, it encounters a trail that is a rare observation which it follows till the end. The bottom image in the figure is the labeling of every location in the map using the topic model that was learned online.

The main contribution of this work is in demonstrating that first, robots can use online topic modeling to learn a visual model of their environment with no supervision; second, by using this topic model they can identify interesting, information rich locations; and finally, do a stochastic gradient ascent in semantic information space to explore the environment, collecting data that improves the topic modeling, resulting in better discriminatory power.

\section{Previous Work}
In the following sections we briefly look at some common variants of the exploration problem.

\subsection{Coverage of Known Environments}
If we have prior knowledge about the world then perhaps the simplest form of exploration is coverage, where the goal is to make the robot pass through every point in the given spatial region of interest. If the space is free of obstacles, then we can simply use a zig-zag path, sometimes known as a boustrophedon path to cover the world. In the case of known obstacles, Choset et al. \cite{Choset1998} proposed boustrophedon cell decomposition of the world such that each cell can be covered by a simple boustrophedon path; then, given this decomposition, a path can be planned through all the cells. This would result in complete coverage. 

Mannadiar and Rekleitis \cite{Mannadiar2010} later proposed splitting some boustrophedon cells so that the robot does not need to move over previously covered cells, resulting in paths guaranteeing optimal coverage. These paths have been extended for use with the general class of non-holonomic robots,  such as aerial vehicles \cite{Xu2011UAVCoverage}.

\subsection{Exploration for Improving Navigation}
Navigating a robot through free space is a fundamental problem in robotics. Yamauchi~\cite{Yamauchi} defined exploration as the ``act of moving through an unknown environment while building a map that can be used for subsequent navigation''. Yamauchi's proposed solution involved moving the robot towards the frontier regions in the map, which were described as the boundary between known free space and the uncharted territories.  

If we have an inverse sensor model of the range sensor, it is possible to compute locations in the world which would maximize the utility of the sensor reading in resolving the obstacle position and shape. Grabowski~\cite{Grabowski} proposed such an exploration strategy in which the goal is to maximize the understanding of obstacles rather than the exposure to free space. In this approach, the robot identifies the location with the next best view, where a sonar sensor reading would have the greatest utility in improving the quality of the representation of an obstacle. 

If there is no external localizer available to the robot, then it is desirable that the robot explores, maps, and localizes in the environment at the same time. Sim, Dudek and Roy \cite{sim2004online} take the approach of finding trajectories at each step that explore new regions while minimizing the localization uncertainty of the robot as it re-enters a previously mapped region. 


Bourgault et al. \cite{Bourgault2002} and Stachniss et al. \cite{Stachniss2005} have proposed an exploration strategy which uses gradient ascent to move the robot towards areas of high entropy which would maximize map information gain, while still keeping the robot localized. 

Kollar et al. \cite{Kollar2008,kollar2008efficient} formulated the exploration problem as a constrained optimization problem, where the goal is to find a path that maximizes map accuracy with the constraint of complete map coverage. To do this, the algorithm first identifies the locations on the map that are essential for coverage, and then uses these locations to constrain the trajectory that maximizes map accuracy. 


\subsection{Exploration for Monitoring Spatiotemporal Phenomena}
In underwater and aerial environments, obstacle avoidance and map building tasks are typically not of primary concern. 

Binney et al. \cite{Binney2013} have described an exploration technique to optimize monitoring spatiotemporal phenomena by taking advantage of the submodularity of the objective function. Bender et al. \cite{Bender2013} has proposed a Gaussian process based exploration technique for benthic environments, which uses an experiment specific utility function. Das et al.~\cite{Das2012} have presented techniques to autonomously observe oceanographic features in the open ocean. Hollinger et al.~\cite{Hollinger2012} have studied the problem of autonomously studying underwater ship hulls by maximizing the accuracy of the sonar data stream. Smith et al.~\cite{Smith2011} have looked at computing robot trajectories which maximize the information gained, while minimizing the deviation from the planned path. 
 
\subsection{Exploration using Topic Modeling}
In our previous work \cite{Girdhar2013IJRR} we used spatiotemporal topic modeling to describe the scene observed by a robot using topic distributions, which acts as a high level scene descriptor that is immune to low level scene changes. We used these descriptors to define an online summary, consisting of a small set of images that are representative of the diversity of the images observed by the robot thus far, and then use these summary images to compute the novelty or surprise score of a newly observed image. This surprise score was used to control the speed of the robot of a pre-defined trajectory. 

The work that we present in this paper improves upon our prior work in many different ways. First, instead of computing novelty of the entire image, we compute the surprise score for different sections of the incoming image observation, which gives us the capability to automatically compute information rich exploration trajectories, and not just control the speed. Second, we use model perplexity to compute the surprise score, instead of the summary based surprise score. Perplexity scores are better suited as surprise score because they have a natural meaning in terms of information gain and uncertainty, and are free of parameters such as summary size. Finally, this work consists of extensive quantitative evaluation of the proposed exploration strategy, and compares it to other exploration strategies. 

\section{Topic Modeling of Observation Data}
In this section we will briefly describe topic modeling process used by ROST \cite{Girdhar2013IJRR}, which we use to give high level labels to the low level features observed by the robot, and also to compute the perplexity score of the observations. 

\subsection{Generative Model}
An observation word is a discrete observation made by a robot. Given the observation words and their location, we would like to compute the posterior distribution of topics at this location. Let $w$ be the observed word at location $x$. We assume the following generative process for the observation words:

\begin{enumerate}
\item word distribution for each topic $k$: $$\phi_k \sim \mathrm{Dirichlet}(\beta),$$
\item topic distribution for words at location $x$ : $$\theta_{x} \sim \mathrm{Dirichlet}(\alpha + H(x)),$$
\item topic label for $w$: $$z \sim \mathrm{Discrete}(\theta_{x}),$$
\item word label: $$w \sim \mathrm{Discrete}(\phi_{z}),$$
\end{enumerate}
where $y\sim Y$ implies that random variable $y$ is sampled from distribution $Y$, $z$ is the topic label for the word observation $w$, and $H(x)$ is the distribution of topics in the neighborhood of location $x$. Each topic is modeled by distribution $\phi_k$ over $V$ possible word in the observation vocabulary. 

\begin{eqnarray}
\phi_k(v) = \Prob(w=v|z=k)  \propto  n^v_k + \beta,
\end{eqnarray}
where $n^v_k$ is the number of times we have observed word $v$ taking topic label $k$, and $\beta$ is the Dirichlet prior hyperparameter. Topic model $\Phi=\{\phi_k\}$ is a $K \times V$ matrix that encodes the global topic description information shared by all locations.

The main difference between this generative process and the generative process of words in a text document as proposed by LDA \cite{Blei:2003, Griffiths:2004} is in step 2. The context of words in LDA is modeled by the topic distribution of the document, which is independent of  other documents in the corpora. We relax this assumption and instead propose the context of an observation word to be defined by the topic distribution of its spatiotemporal neighborhood. This is achieved via the use of a kernel. The posterior topic distribution at location $x$ is thus defined as: 
\begin{eqnarray}
\theta_x(k) = \Prob(z=k|x) \propto \left( \sum_y K(x-y) n^k_y\right) + \alpha,
\end{eqnarray}
where $K(\cdot)$ is the kernel, $\alpha$ is the Dirichlet prior hyperameter and, $n^k_y$ is the number of times we observed topic $k$ at location $y$.

\subsection{Approximating Neighborhoods using Cells}
The generative process defined above models the clustering behavior of observations from a natural scene well, but is difficult to implement because it requires keeping track of the topic distribution at every location in the world. This is computationally infeasible for any large dataset. For the special case when the kernel is a uniform distribution over a finite region, we can assume a cell decomposition of the world, and approximate the topic distribution around a location by summing over topic distribution of cells in and around the location.

Let the world be decomposed into $C$ cells, in which each cell $c\in C$ is connected to its neighboring cells $G(c)~\subseteq~C$. Let $c(x)$ be the cell that contains points $x$. In this paper we only experiment with a grid decomposition of the world in which each cell is connected to its six nearest neighbors, 4 spatial and 2 temporal. However, the general ideas presented here are applicable to any other topological decomposition of the spacetime. Six neighbors is the smallest number which we need to consider while working with streaming 2D image data. 


\begin{figure}[t]
\begin{center}
\includegraphics[width=0.6\columnwidth]{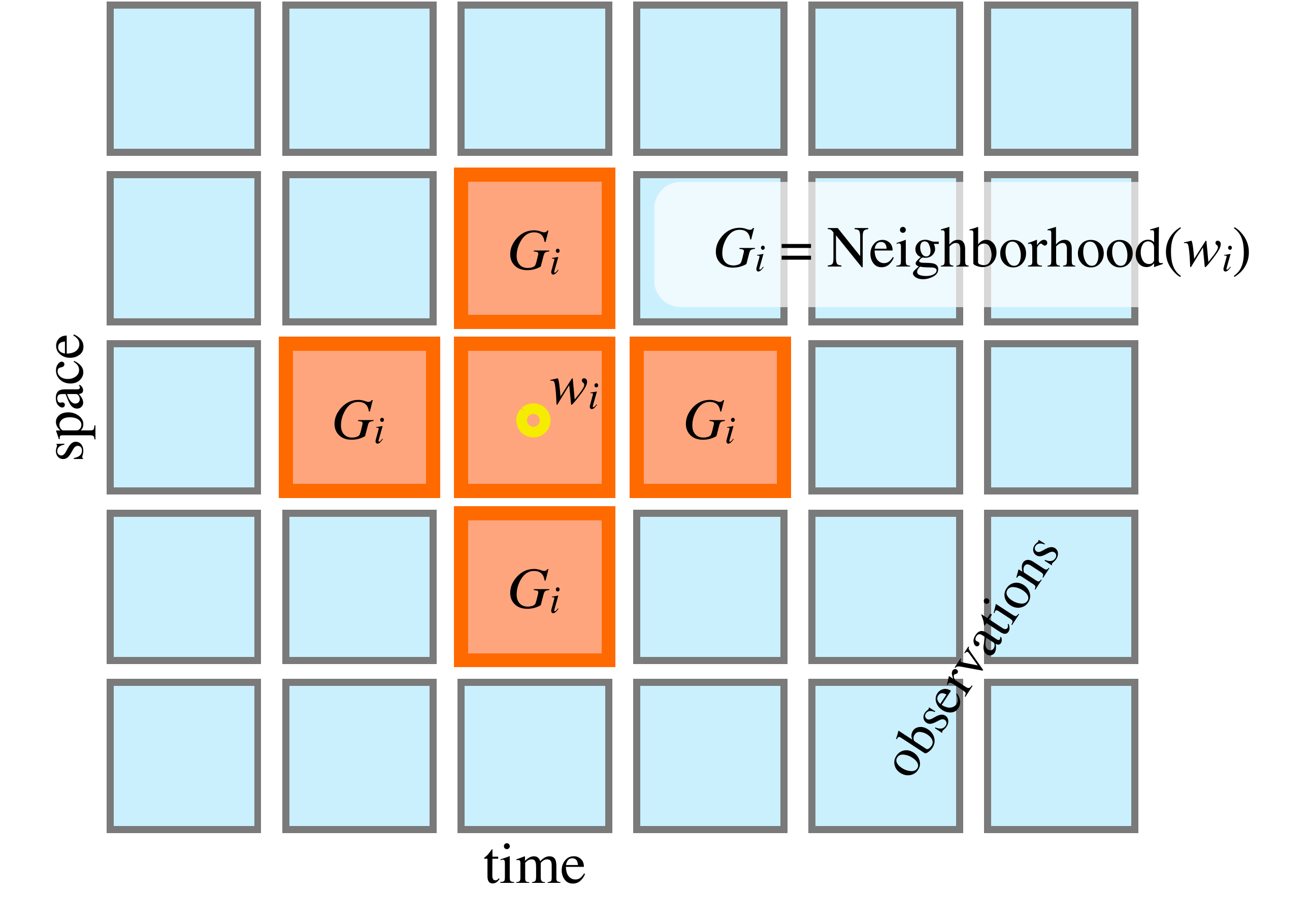}
\caption[Spatiotemporal neighborhood]{Each cell shown corresponds to a spatiotemporal bucket containing all the observation from that region. We refine the topic label for a word $w_i$ in an observation by taking into account the spatiotemporal context $G_i$ of the observation. }\label{fig:neighborhood}

\end{center}
\end{figure}

The topic distribution around $x$ can then be approximated using cells as: 
\begin{eqnarray}
\theta_x(k) \propto \left(\sum_{c' \in G(c(x))} n_{c'}^k\right) + \alpha
\end{eqnarray}

Due to this approximation, the following properties emerge:
\begin{enumerate}
\item $\theta_x = \theta_y$ if $c(x) = c(y)$, i.e., all the points in a cell share the same neighborhood topic distribution.
\item The topic distribution of the neighborhood is computed by summing over the topic distribution of the neighboring cells rather than individual points. 
\end{enumerate}
We take advantage of these properties while doing inference in realtime.

\subsection{Realtime Inference using Gibbs Sampling}

Given a word observation $w_i$, its location $x_i$, and its neighborhood $G_i = G(c(x_i))$, we use a Gibbs sampler to assign a new topic label to the word, by sampling from the posterior topic distribution:
\begin{eqnarray}
\begin{split}
\Prob(z_i = k |  w_i=v,x_i) \propto  \frac{n_{k,-i}^{v} + \beta}{\sum_{v=1}^V (n_{k,-i}^{v} + \beta)} \cdot \\
\frac{n_{G_i,-i}^{k} + \alpha}{\sum_{k=1}^K (n_{G_i,-i}^{k} + \alpha)},
\end{split} \label{eq:refine}
\end{eqnarray}
where $n_{k,-i}^{w}$ counts the number of words of type $w$ in topic $k$, excluding the current word $w_i$, $n_{G_i,-i}^{k}$ is the number of words with topic label $k$ in neighborhood $G_i$, excluding the current word $w_i$, and $\alpha, \beta$ are the Dirichlet hyper-parameters.  Note that for a neighborhood size of 0, the above Gibbs sampler is equivalent to the LDA Gibbs sampler proposed by Griffiths et al.\cite{Griffiths:2004}, where each cell corresponds to a document. Algorithm \ref{alg:batchgibbs} shows a simple iterative technique to compute the topic labels for the observed words in batch mode. 

\begin{algorithm}[t]	
	\DontPrintSemicolon
	\caption{Batch Gibbs sampling} 
	\label{alg:batchgibbs}
    Initialize $ \forall i, z_i \sim ~ \mathrm{Uniform}(\{1,\dots,K\})$ \;
    
	\While{true}{
  		\ForEach{ cell $c \in C$ }{   
	  		\ForEach{ word $w_i \in c $ }{   
	  			$z_i \sim \Prob(z_i = k |  w_i=v, x_i)$ \;
	  			Update $\Theta, \Phi$ given the new $z_i$ by updating $n_k^v$ and $n_G^k$\;
	  		}
        }	
	}
\end{algorithm}

In the context of robotics we are interested in the online refinement of observation data. After each new observation, we only have a constant amount of time to do topic label refinement. Hence, any online refinement algorithm that has computational complexity which increases with new data, is not useful. Moreover, if we are to use the topic labels of an incoming observation for making realtime decisions, then it is essential that the topic labels for the last observation converge before the next observation arrives. 

Since the total amount of data collected grows linearly with time, we must use a refinement strategy that efficiently handles global (previously observed) data and local (recently observed) data. 

Our general strategy is described by Algorithm \ref{alg:rostgibbs}. At each time step we add the new observations to the model, and then randomly pick observation times $t\sim\Prob(t|T)$, where $T$ is the current time, for which we resample the topic labels and update the topic model. 

\begin{algorithm}[t]	
	\DontPrintSemicolon
	\caption{Realtime Gibbs sampler} 
	\label{alg:rostgibbs} \label{alg:refinetopics}
	\While{true}{
	     Add new observed words to their corresponding cells.\;
        Initialize $ \forall i\in M_T, z_i \sim ~ \mathrm{Uniform}(\{1,\dots,K\})$ \;
    
	    \While{no new observation}{
	        $t \sim \Prob(t|T)$\;
            \ForEach{ cell $c \in M_t$ }{   
			      \ForEach{ word $w_i \in c $ }{   
				    $z_i \sim \Prob(z_i = k |  w_i=v, x_i)$ \;
				    Update $\Theta, \Phi$ given the new $z_i$ by updating $n_k^v$ and $n_G^k$\;
			       }
            }	
        }
        $T \leftarrow T+1$\;	            
	}
\end{algorithm}

We choose $\Prob(t|T)$ such that with probability $\eta$ we refine the last observation, and with probability $(1-\eta)$ we refine a randomly picked previous observation. We call $\eta$ the refinement bias of the Gibbs sampler. 

\begin{eqnarray}
\Prob(t|T) = 
\begin{cases}
\eta,& \text{if }t=T \\
(1-\eta)/(T-1),& \text{otherwise} 
\end{cases}
\end{eqnarray}

\section{Curiosity based Exploration} 
We assume a cellular decomposition of the world, in which each cell $c\in C$ is connected to its neighboring cells $G(c)~\subset~C$. The world is composed of at most $K$ different kinds of terrains or other high level visual objects (which we refer to as topics), each of which, when observed by a robot, can result in $V$ different kinds of low level observations, where $V >> K$. Each topic $k$ is described by a distribution $\phi_k$ over these $V$ different types of observations, and for any cell $c$, $\phi_{G(c)}$ is the distribution of topics in and around the cell. The goal then is to plan a continuous path $P \subseteq C$, that allows us to learn the topic model $\Phi=\{\phi_k\}$ that best describes the world by labeling each observation at each location with a representative topic label. 

At time $t$, let the robot be in cell $p_t=c$, and let $G(c)=\{g_i\}$ be the set of cells in its neighborhood. We would like to compute a weight value for each $g_i$, such that the probability of the robot taking a step in this direction is proportional to this weight. 
\begin{eqnarray}
\Prob(p_{t+1} = g_i) \propto \mathrm{weight}(g_i).
\end{eqnarray}

In this work we consider four different weight functions, one that is completely unaware of of its surrounding, one that is only spatially aware and tries to cover the unexplored free space, and two that are both spatially and observationally aware.
\begin{enumerate}
\item Random Walk - Each cell in the neighborhood is equally likely to be the next step:
    \begin{eqnarray}
    \mathrm{weight}(g_i) = 1.
    \end{eqnarray}
\item Stochastic Coverage - Use a potential function to repel previously visited locations:
    \begin{eqnarray}
    \mathrm{weight}(g_i) = \frac{1}{\sum_j n_j/d^2(g_i,c_j)}.
    \end{eqnarray}
    where $n_j$ is the number of times we have visited cell $c_j$, and $d(g_i,c_j)$ is the Euclidean distance between these two cells. 

\item Word Perplexity - Bias the next step towards cells which have high word perplexity:
    \begin{eqnarray}
    \mathrm{weight}(g_i) = \frac{\mathrm{WordPerplexity}(g_i)}{\sum_j n_j/d^2(g_i,c_j)}.
    \end{eqnarray}

\item Topic Perplexity - Bias the next step towards cells which have high topic perplexity:
    \begin{eqnarray}
    \mathrm{weight}(g_i) = \frac{\mathrm{TopicPerplexity}(g_i)}{\sum_j n_j/d^2(g_i,c_j)}.
    \end{eqnarray}
    
\end{enumerate}

We compute the word perplexity of the words observed in $g_i$ by taking the inverse geometric mean of the probability of observing the words in the cell, given the current topic model and the topic distribution of the path thus far. 
\begin{eqnarray}
\begin{split}
\mathrm{WordPerplexity}(g_i) = \exp\left(-\frac{\sum_i^{W} \log  \sum_k \Prob(w_i=v | k)\Prob(k | P)  }{W} \right),
\end{split}
\end{eqnarray}
where $W$ is the number of words observed in $g_i$, $\Prob(w_i=v|k)$ is the probability of observing word $v$ if its topic label is $k$, and $\Prob(k|P)$ is the probability of seeing topic label $k$ in the path executed by the robot thus far.

To compute topic perplexity of the words observed in $g_i$, we first compute topic labels $z_i$ for these observed words by sampling them from the distribution in Eq. \ref{eq:refine}, without adding these words to the topic model. These temporary topic labels are then used to compute the perplexity of $g_i$ in topic space.

\begin{eqnarray}
\begin{split}
\mathrm{TopicPerplexity}(g_i) = \exp\left(-\frac{\sum_i^{W} \log \Prob(z_i=k | P)  }{W} \right).
\end{split}
\end{eqnarray}

Note that due to presence of repulsive potential from the previously visited location, and stochastic nature of how the next step is taken, the robot is unlikely to get caught in a local maxima.

\section{Experiments}

\begin{figure}[]
\begin{center}

\subfigure[]{\includegraphics[width=0.32\columnwidth]{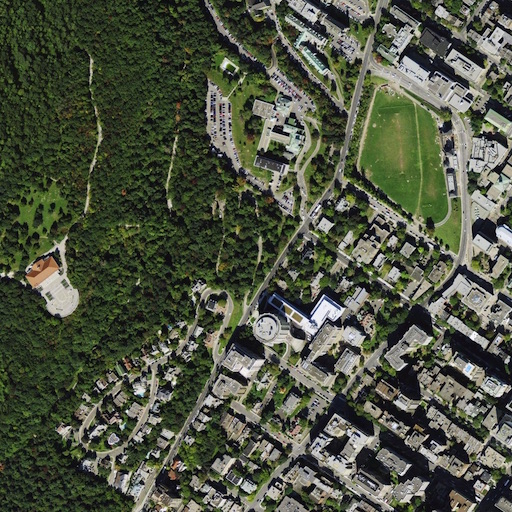}}
\subfigure[]{\includegraphics[width=0.32\columnwidth]{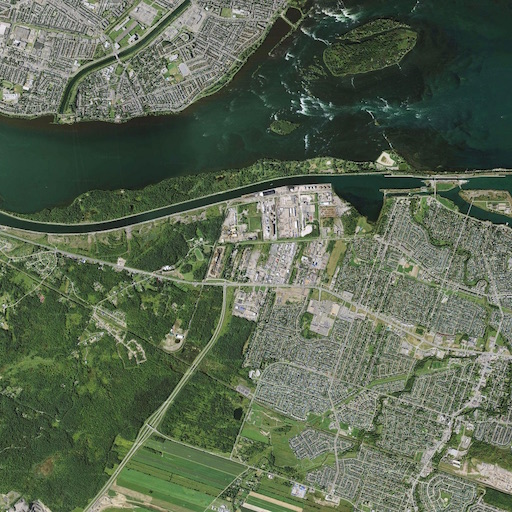}}
\subfigure[]{\includegraphics[width=0.32\columnwidth]{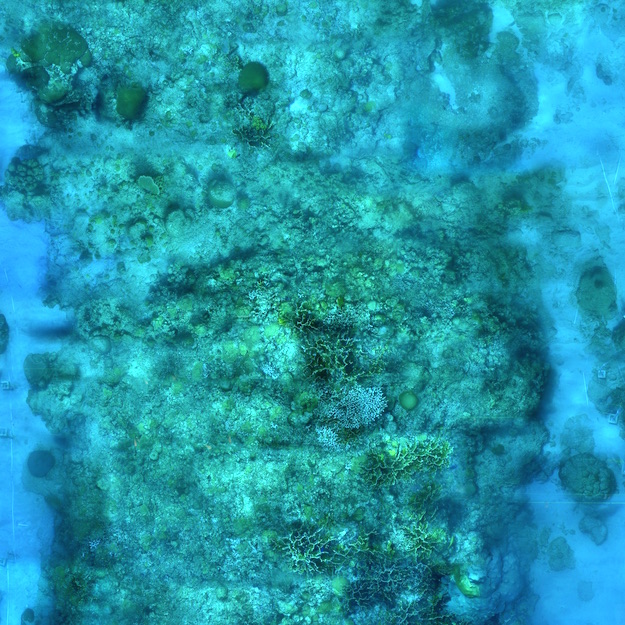}}

\subfigure[]{\includegraphics[width=0.32\columnwidth]{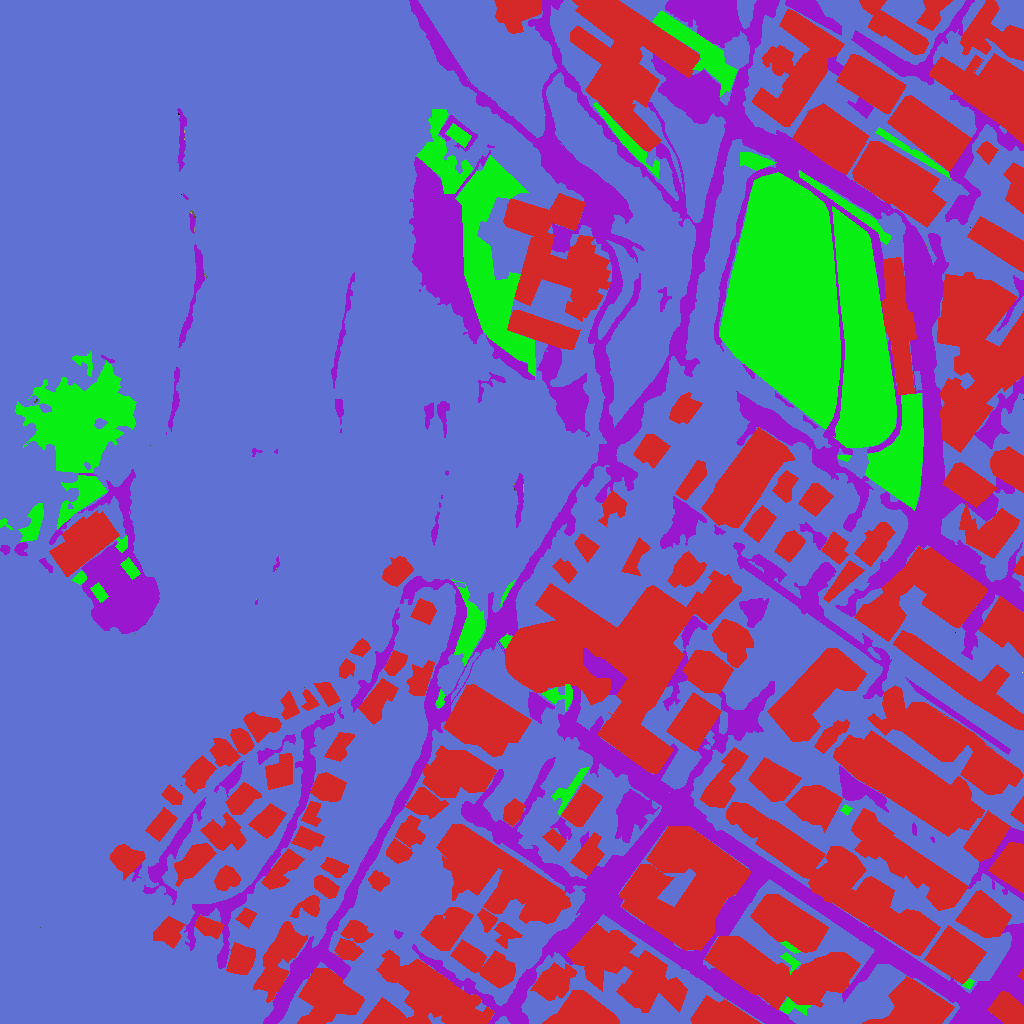}}
\subfigure[]{\includegraphics[width=0.32\columnwidth]{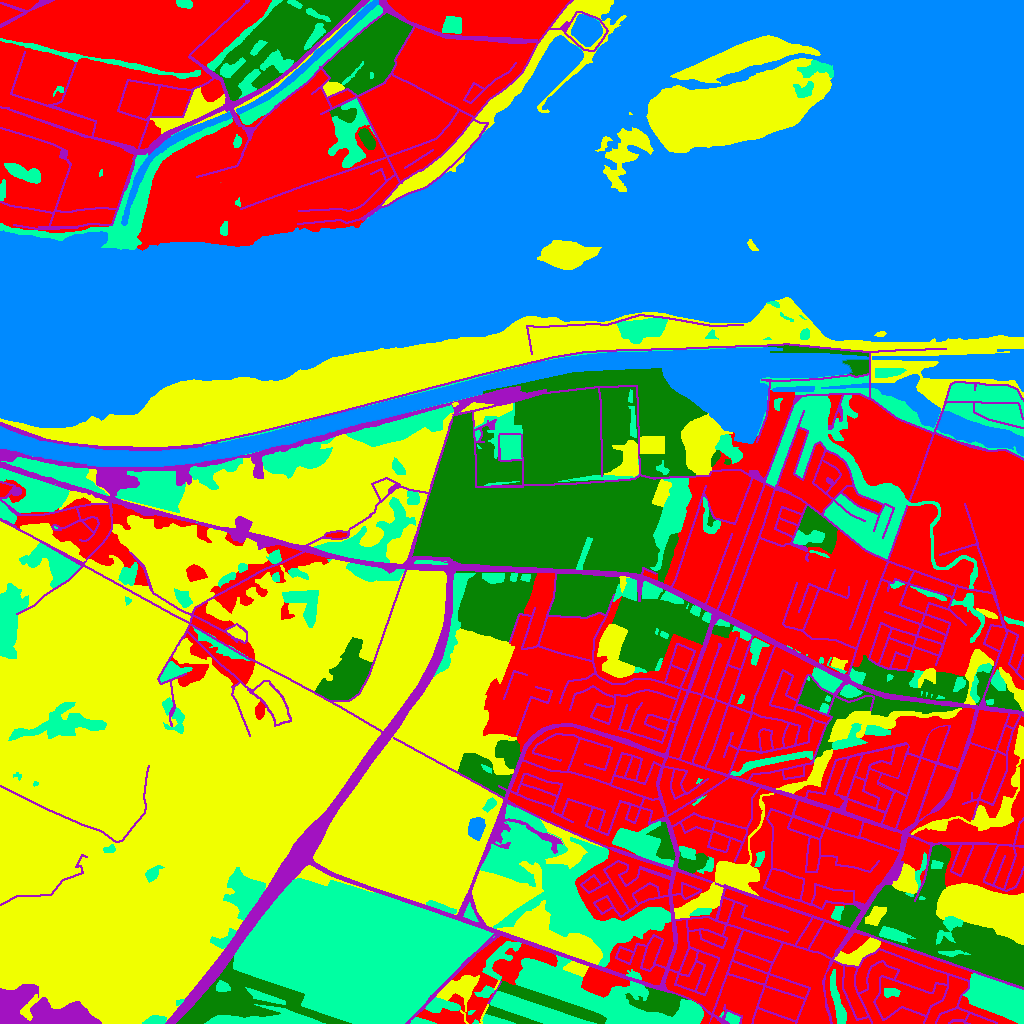}}
\subfigure[]{\includegraphics[width=0.32\columnwidth]{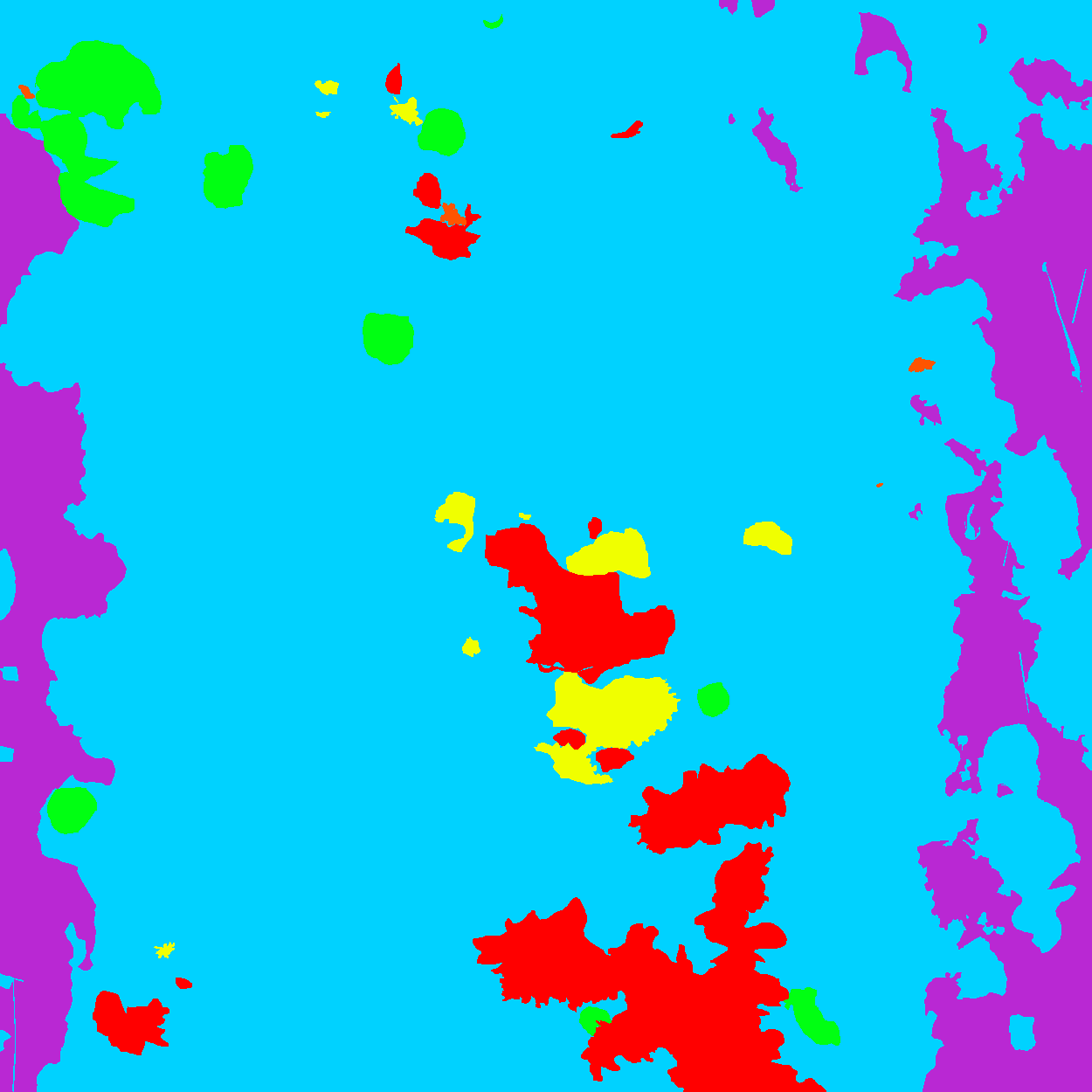}}


\subfigure[]{\includegraphics[width=0.32\columnwidth]{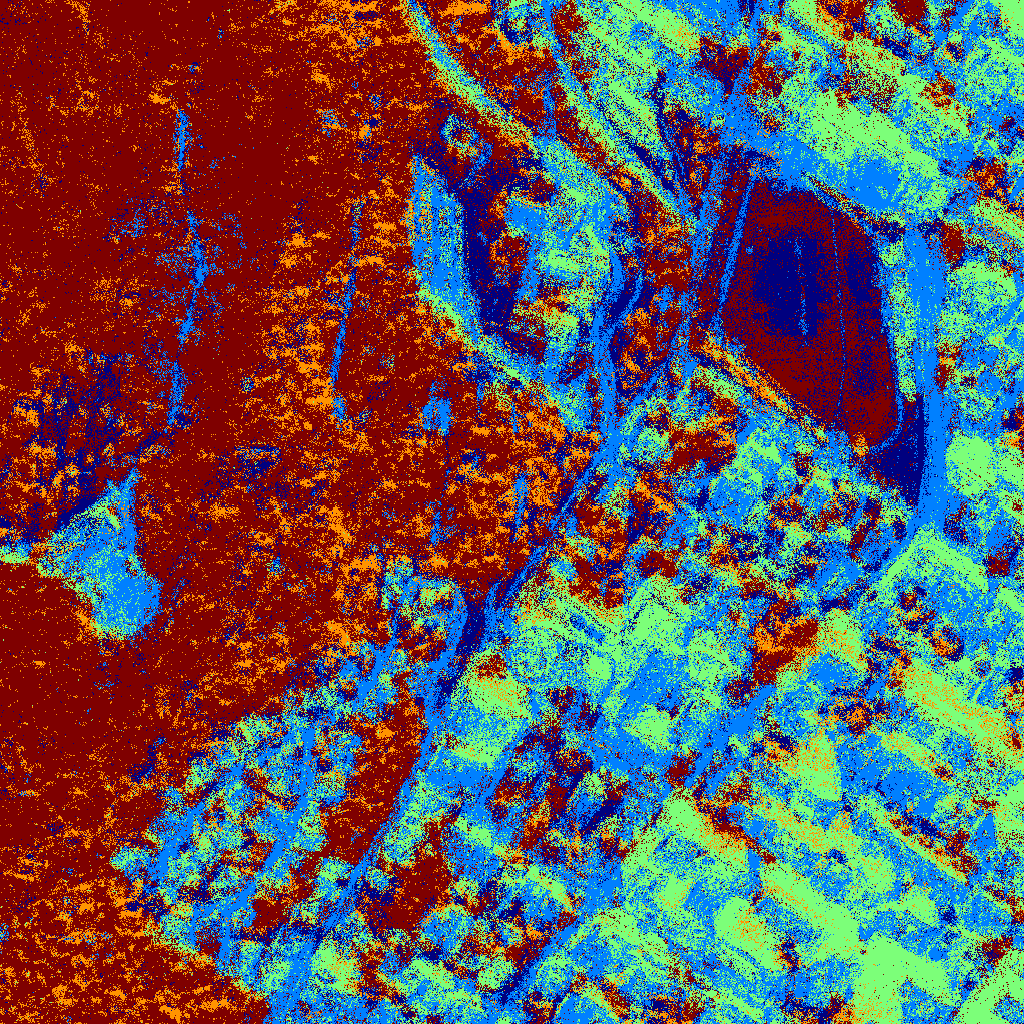}}
\subfigure[]{\includegraphics[width=0.32\columnwidth]{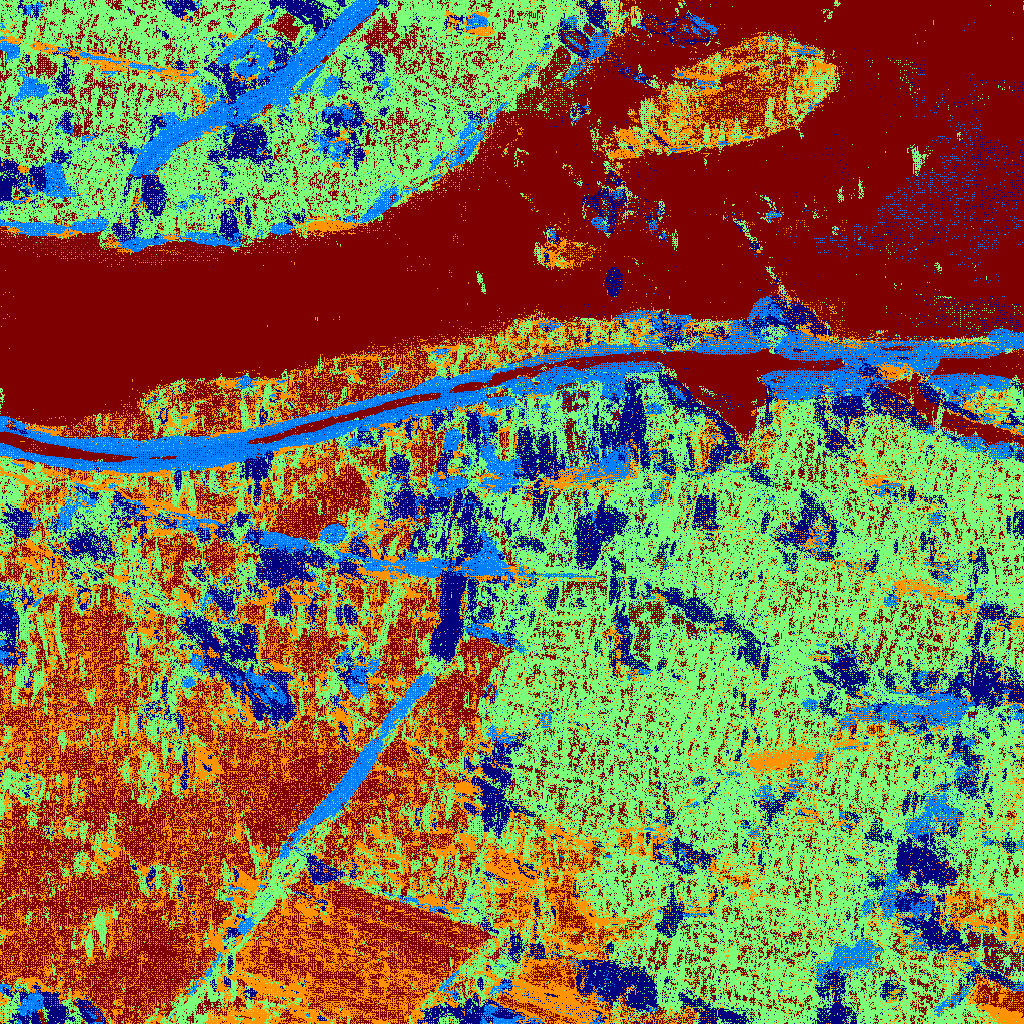}}
\subfigure[]{\includegraphics[width=0.32\columnwidth]{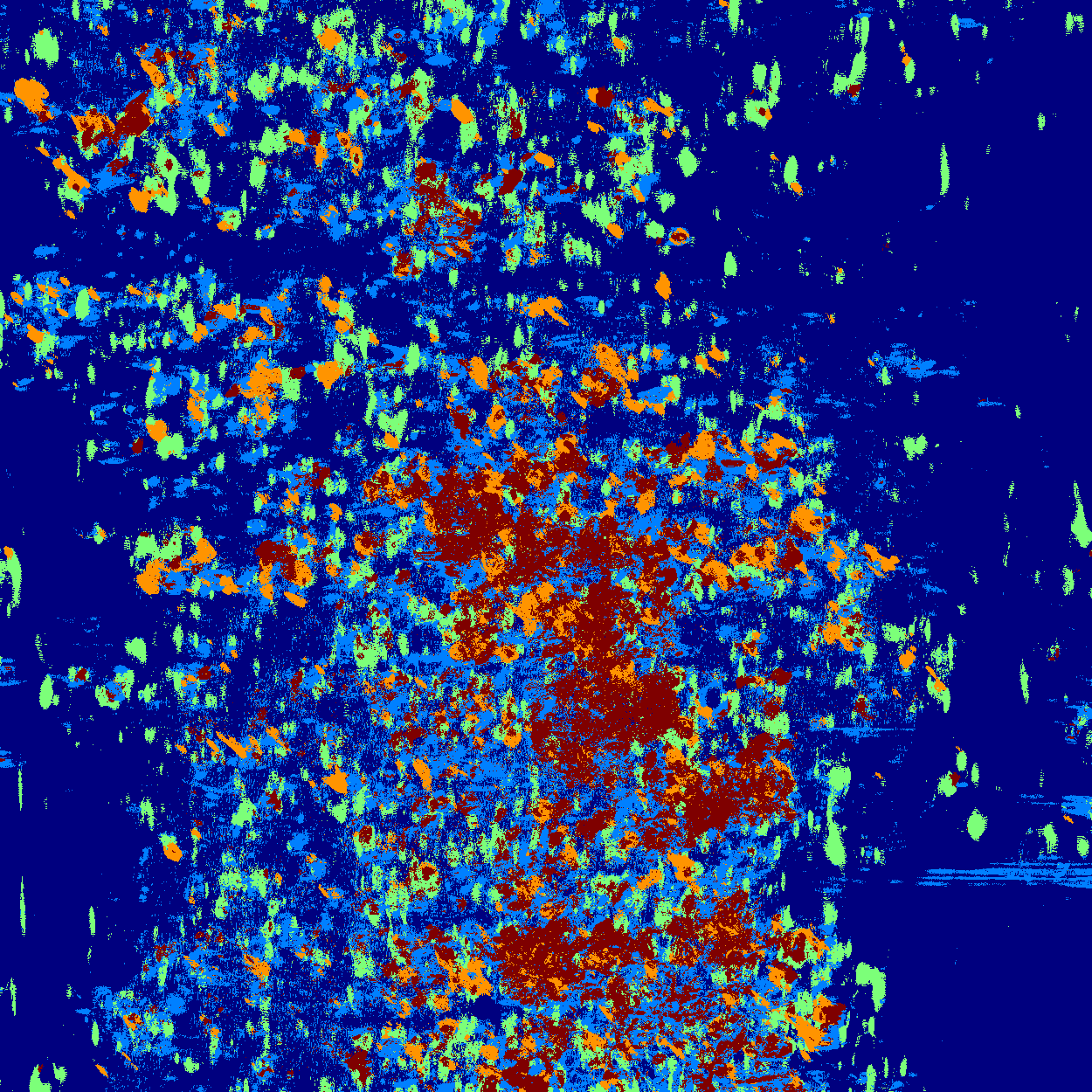}}

\caption[Example of results of curiosity based exploration on a 2D dataset]{Example of results of curiosity based exploration on a 2D dataset. (a)-(c) Input image used to generate observation data, (d)-(f) Groundtruth labeling. 
(g)-(i) Terrain labeling of the map using the topic model computed on the path. }\label{fig:ds}
\end{center}
\end{figure}

\begin{figure}[]
\begin{center}

\includegraphics[width=0.04\columnwidth]{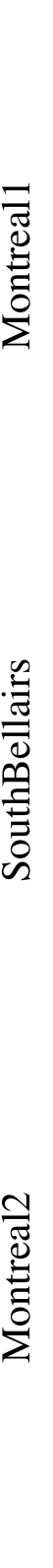} 
\includegraphics[width=0.9\columnwidth]{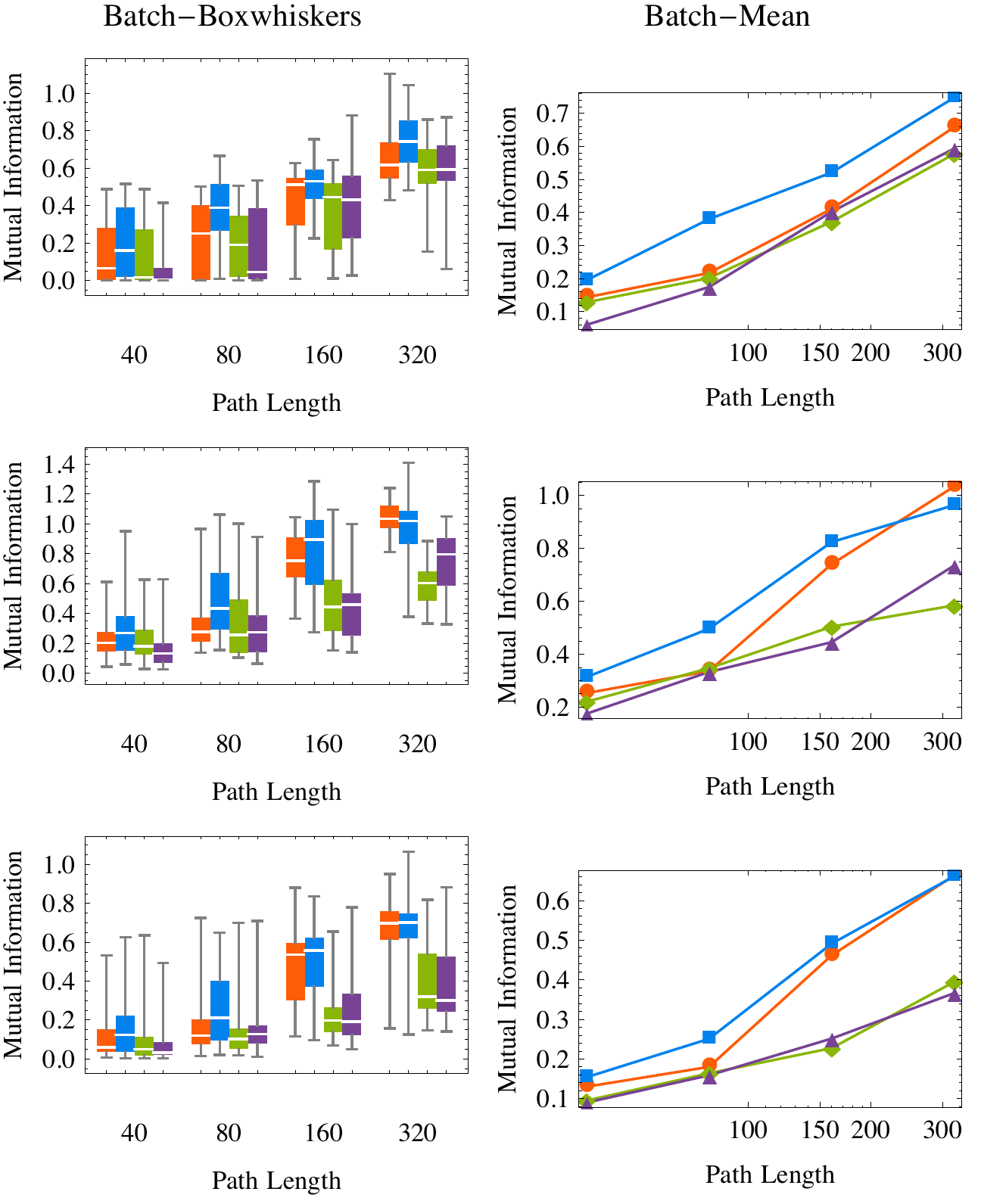} \\
\includegraphics[width=0.2\columnwidth]{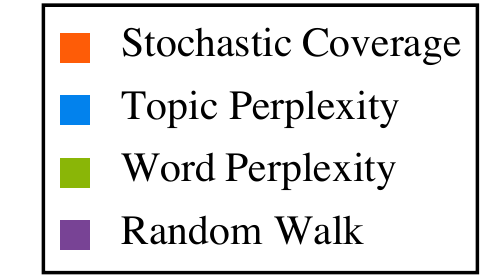} 
\caption[Evaluating exploration on a 2D map]{Evaluation of the proposed exploration techniques. The plots show mutual information between the maps labeling produced using the topic model computed online during the exploration, with maps labeled by batch processing of the data.}\label{fig:plots2dexplore:batch}
\end{center}
\end{figure}

\begin{figure}[]
\begin{center}
\includegraphics[width=0.04\columnwidth]{mutual_info_plot_names} 
\includegraphics[width=0.9\columnwidth]{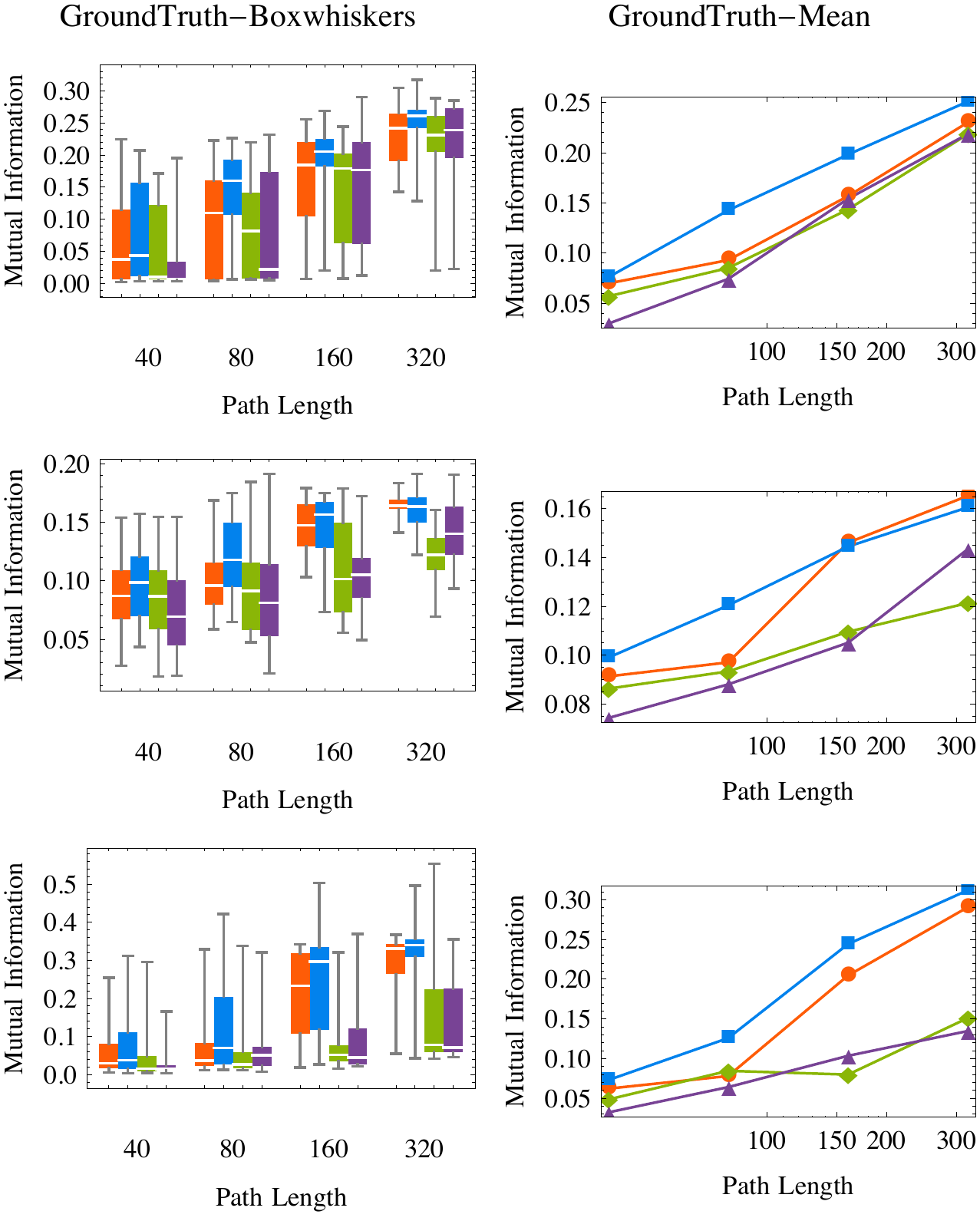} \\
\includegraphics[width=0.2\columnwidth]{mi_plot_legend} 
\caption[Evaluating exploration on a 2D map]{Evaluation of the proposed exploration techniques. The plots show mutual information between the maps labeling produced using the topic model computed online during the exploration, with maps labeled by a human}\label{fig:plots2dexplore:gt}
\end{center}
\end{figure}

\subsection{Exploration on a 2D Map}
\subsubsection{Setup}
To validate our hypothesis that biasing exploration towards high perplexity cells will result in a better terrain topic model of the environment, we conducted the following experiment. We considered three different maps: two aerial views, and one underwater coral reef map.  
\begin{table}
\begin{center}
\begin{tabular}{|l|l|l|l|l|}
\hline
Dataset & width(px) & height(px) & n.cells & n.words\\
\hline
Montreal1 (aerial) & 1024 & 1024 & 4096 & 3,239,631\\
Montreal2 (aerial) & 1024 & 1024 & 4096 & 1,675,171\\
SouthBellairs (underwater) & 2500 & 2500 & 6241 & 1,664,749\\
\hline
\end{tabular}
\caption{Exploration dataset specifications} \label{tab:exploration_ds}
\end{center}
\end{table}

We extracted ORB words describing local features, and texton words describing texture at every pixel (every second pixel for the SouthBellairs underwater dataset). ORB~\cite{RubleeE2011} words had a dictionary size of 5000, and texton words had a dictionary size of 1000. The dictionary was computed by extracting features from a completely unrelated dataset.

Each of these maps were decomposed into square cells of width 16 pixels (32 for SouthBellairs). Now for each weight function, we computed exploration paths of varying length, with 20 different random restart locations for each case. Each time step was fixed at 200 milliseconds to allow the topic model to converge. We limited the path length to 320 steps, which is about $5\sqrt|C|$. Some basic statistics about the three datasets are given in Table \ref{tab:exploration_ds}

Each of these exploration runs returned a topic model $\Phi_p$, which we then used to compute topic labels for each pixel in the map in batch mode. Let $Z_p$ be these topic labels. An example of this labeling for each of the three dataset is shown in the last row of Figure \ref{fig:ds}. We compared this topic labeling with two other labelings: human labeled ground-truth $Z_h$, and labels computed automatically in batch mode $Z_b$, where we assume random access to the entire map. 


We then computed the mutual information between $Z_p$ and $Z_h$, 
$Z_p$ and $Z_b$, and plotted the results as a function of path length, as shown in Figure \ref{fig:plots2dexplore:batch} and \ref{fig:plots2dexplore:gt}.

\subsubsection{Results}\label{sec:results2dexplore}
The results are both encouraging and surprising. As shown in Figure \ref{fig:plots2dexplore:batch} and\ref{fig:plots2dexplore:gt}, we see that topic perplexity based exploration (shown with blue squares) performs consistently better than all other weight functions, when compared against ground truth, or the batch results. 

For paths of length 80, which is close to the width of the maps, we see that mutual information between  topic perplexity based exploration and ground truth is 1.51, 1.20 and 1.05  times higher respectively for the three datasets, compared to the next best performing technique. 

For long path lengths (320 steps or more), stochastic coverage (shown with orange circles) based exploration matches the mean performance of topic perplexity exploration. This is expected because the maps are bounded, and as the path length increases, the stochastic coverage algorithm is able to stumble across different terrains, even without a guiding function. 

For short path lengths (40 steps or less), we do not see any statistical difference between the performance of different techniques.

Marked with purple triangles, we see the results of exploration using Brownian random motion.  Although this strategy has a probabilistic guarantee of asymptotically complete coverage, but it does so at a lower rate that stochastic coverage exploration strategy.
A random walk in two dimensions is expected to travel a distance of $\sqrt n$ from start, where $n$ is the number of steps. Hence it is highly likely that it never visits different terrains. The resulting topic models from these paths are hence unable to resolve between these unseen terrains. 

The performance of word perplexity exploration (shown with green diamonds) is surprisingly poor in most cases. We hypothesize that this poor performance is due to the algorithm getting pulled towards locations with terrain described by a more complex word distribution. This will cause the algorithm to stay in these complex terrains, and not explore as much as the other algorithms. In comparison, the topic perplexity exploration is not affected by the complexity of the distribution describing the topic, and is only attracted to topic rarity.

\subsection{Demonstration: Underwater Exploration}
\begin{figure}[]
\begin{center}

\subfigure[]{\includegraphics[clip=true, trim = 57 80 57 34, width=0.48\columnwidth]{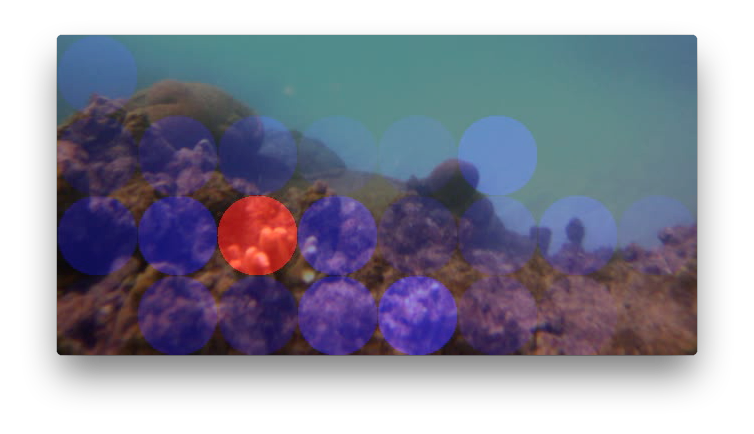}}
\subfigure[]{\includegraphics[clip=true, trim = 57 80 57 34, width=0.48\columnwidth]{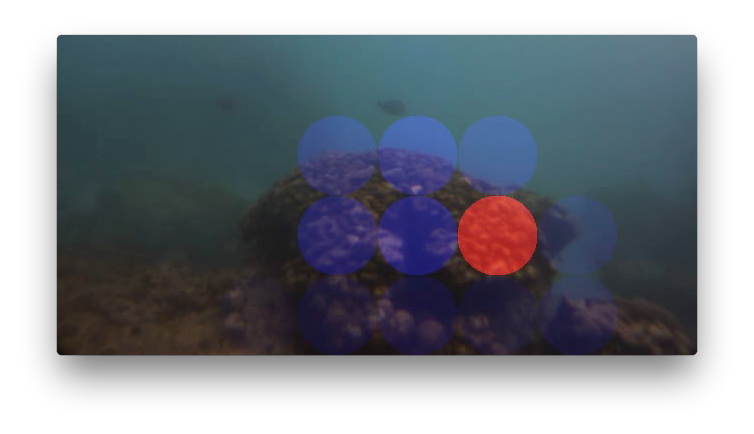}}
\subfigure[]{\includegraphics[clip=true, trim = 57 80 57 34, width=0.48\columnwidth]{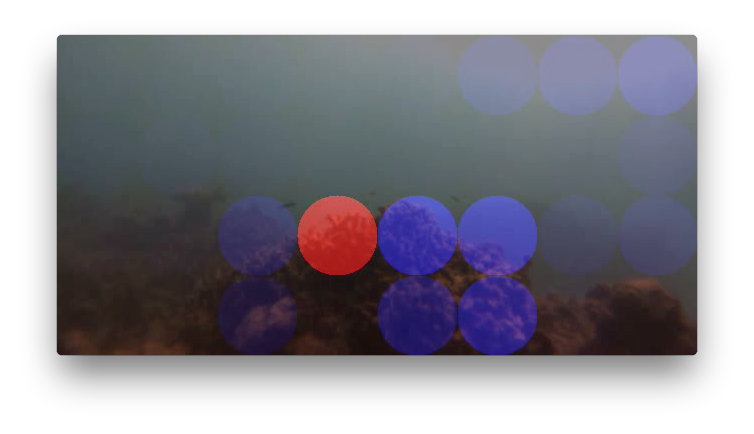}}
\subfigure[]{\includegraphics[width=0.48\columnwidth]{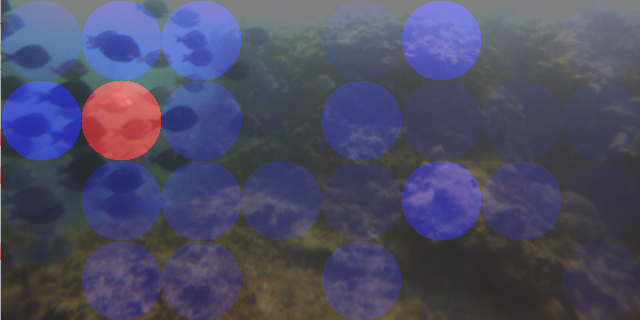}}
\subfigure[]{\includegraphics[clip=true, trim = 57 80 57 34, width=0.48\columnwidth]{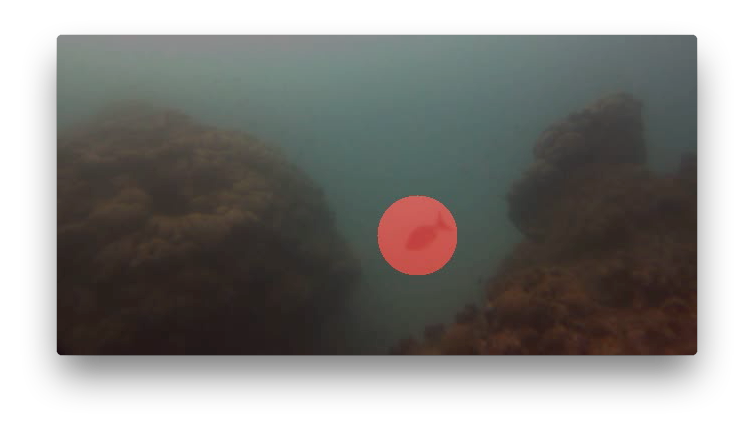}}
\subfigure[]{\includegraphics[clip=true, trim = 57 80 57 34, width=0.48\columnwidth]{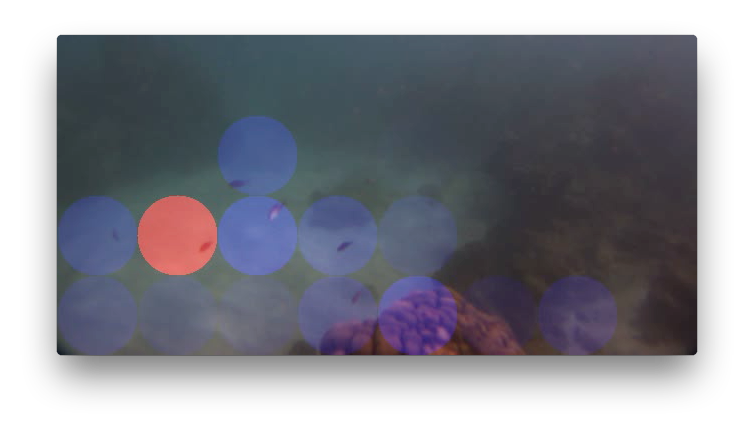}}
\subfigure[]{\includegraphics[clip=true, trim = 57 80 57 34, width=0.48\columnwidth]{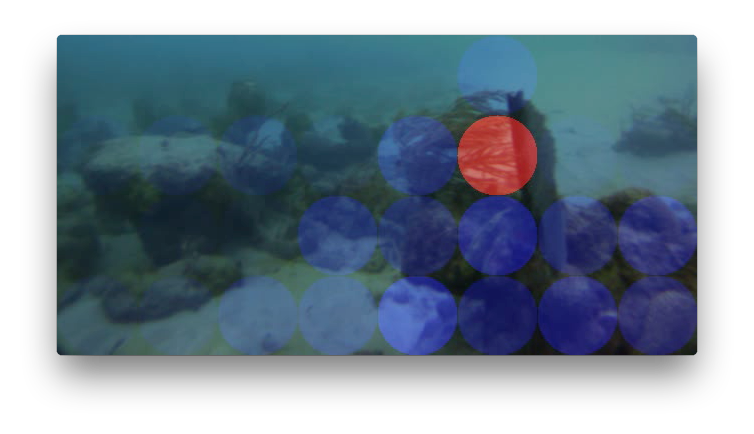}}
\subfigure[]{\includegraphics[clip=true, trim = 57 80 57 34, width=0.48\columnwidth]{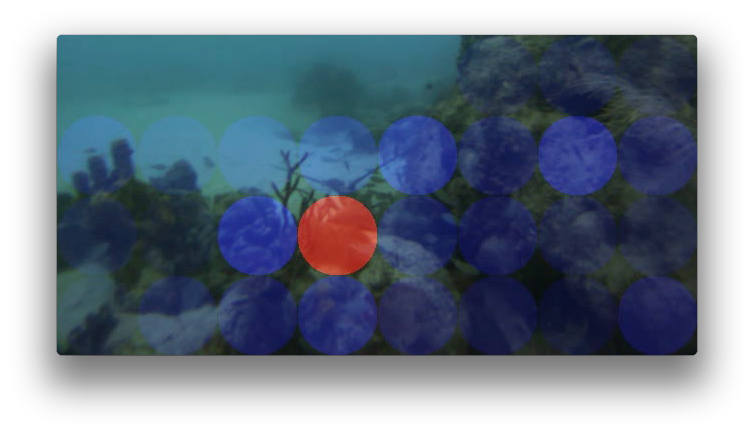}}
\subfigure[]{\includegraphics[clip=true, trim = 57 80 57 34, width=0.48\columnwidth]{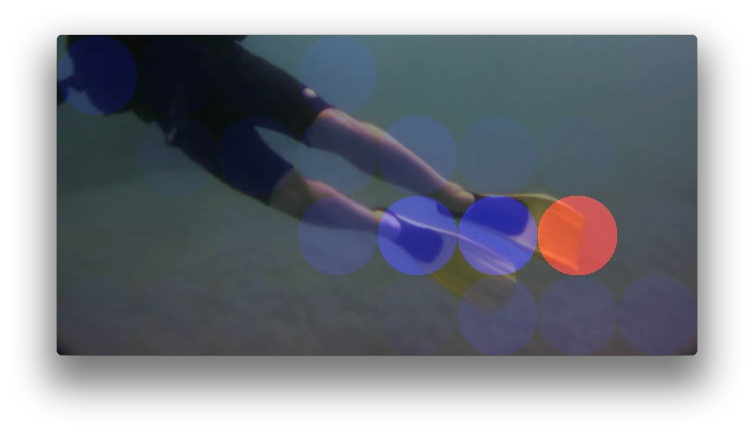}}
\subfigure[]{\includegraphics[width=0.48\columnwidth]{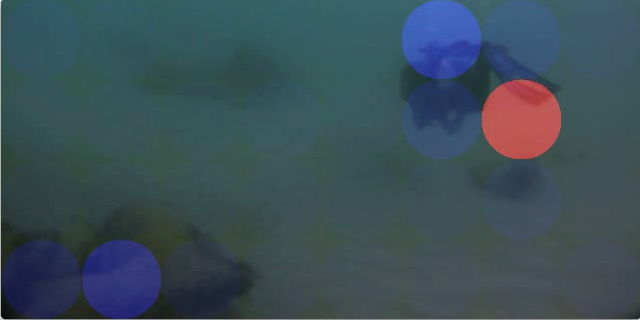}}

\caption[Examples of high perplexity patches in underwater scenes]{Examples of observations showing cells marked with their curiosity score. Red marks the cell with the highest score. }\label{fig:uw_curiosity}

\end{center}
\end{figure}

We implemented the proposed curiosity modeling system on Aqua amphibious robot \cite{Dudek:Comp:2007,Sattar:IROS:2008}, and tested it in three different underwater scenarios as shown in the video located at: \url{http://cim.mcgill.ca/mrl/girdhar/rost/aqua_curiosity.mp4}. In this video we see the robot exploring its environment from two different points of view. We color the cells in robot's view with blue, and change the opacity based on the perplexity score. A cell marked with more opaque blue circle has higher topic perplexity score, and the cell with the highest score is marked with a red color. Figure \ref{fig:uw_curiosity} shows some examples of these high perplexity regions in observed images by the robot.  For all our experiments, we fixed the number of topics to $K=64$, and set Dirichlet hyper-parameters $\alpha=0.1$, $\beta=0.1$, refinement bias $\eta = 0.5$, and cell curiosity score decay rate of $\gamma = 0.7$.

\subsubsection{Scenario 1: Exploring a coral head} 
In this trial, we started the robot near a coral head surrounded by monotonous sand. We see that the robot quickly gets attracted towards the coral head, and continues to bounce around over this structure while staying away from sand. We see the effect of curiosity decay variable $\gamma$, as the robot is successfully able to return back to the coral head several times after going over the much less interesting sandy regions.

\subsubsection{Scenario 2: Interaction with a diver}
Although our goal was to study the robot as it would interact with a fish, due to lack of cooperation with the fish, we were forced to conduct the experiment with a scuba diver instead. We see that as soon as the diver is in robot's view, it is the singular source of curiosity for the robot. We see the robot following the diver around, and hovering over the diver when he has stopped moving. 

\subsubsection{Scenario 3: Exploring the ocean floor}
In this trial, we started the robot near the ocean floor, which was sparsely populated with sea plants and corals.  We see the robot manages to keep its focus on sea life, while not wasting time over sand.

\section{Summary} 
In this paper we have presented a long-term exploration technique that aims to learn a observation model of the world by finding paths with high information content. The use of a realtime, life-long learning, topic modeling framework allows us to describe the incoming streams of low level observation data via the use of latent variables representing the terrain type. Given this online, life-long learning model, we compute the utility of the potential next steps in the path in terms of their perplexity scores. We validated the effectiveness of the proposed exploration technique over candidate techniques by computing mutual information between the terrain maps generated through the use of the learned terrain model, and hand labeled ground truth, on three different datasets.

In our underwater video demonstration, we see that the emergent behavior  of the robot has a striking similarity to that of biological organisms.  While the current work on automated exploration was not explicitly bio-inspired, the relationship between exploration by living agents and the behavior that emerges from this algorithm might be a fruitful direction for further research.

\begin{acknowledgements}
\small{This work was supported by the Natural Sciences and Engineering Research Council (NSERC) through the NSERC Canadian Field Robotics Network (NCFRN). Yogesh Girdhar is currently supported by the Postdoctoral Scholar Program at the Woods Hole Oceanographic Institution, with funding provided by the Devonshire Foundation and the J. Seward Johnson Fund, and  FQRTN Postdoctoral Fellowship. Authors would like to thank Julian Straub for helpful discussion; Philippe Giguere, Ioannis Rekleitis, Florian Shkurti, and Juan Camillio Gamboa for help in conducting field trials.}

\end{acknowledgements}

\bibliographystyle{plain}
\bibliography{library}

\end{document}